# Conditions for Open-Ended Evolution in Immigration Games

Peter D. Turney[1]

[1]Ronin Institute, 127 Haddon Place, Montclair, NJ 07043-2314, USA
peter.turney@roninstitute.org

**Abstract**

The *Immigration Game* (invented by Don Woods in 1971) extends the solitaire *Game of Life* (invented by John Conway in 1970) to enable two-player competition. The Immigration Game can be used in a model of evolution by natural selection, where fitness is measured with competitions. The rules for the Game of Life belong to the family of *semitotalistic rules*, a family with 262,144 members. Woods' method for converting the Game of Life into a two-player game generalizes to 8,192 members of the family of semitotalistic rules. In this paper, we call the original Immigration Game the *Life Immigration Game* and we call the 8,192 generalizations *Immigration Games* (including the Life Immigration Game). The question we examine here is, *what are the conditions for one of the 8,192 Immigration Games to be suitable for modeling open-ended evolution?* Our focus here is specifically on conditions for the rules, as opposed to conditions for other aspects of the model of evolution. In previous work, it was conjectured that *Turing-completeness* of the rules for the Game of Life may have been necessary for the success of evolution using the Life Immigration Game. Here we present evidence that Turing-completeness is a *sufficient* condition on the rules of Immigration Games, but not a *necessary* condition. The evidence suggests that a necessary and sufficient condition on the rules of Immigration Games, for open-ended evolution, is that the rules should allow growth.

## Introduction

The *Game of Life* is a cellular automaton that displays lifelike behaviours. It was invented by John Conway and presented in *Scientific American* by Martin Gardner (Gardner, 1970). The game is played on a potentially infinite, two-dimensional grid of square cells. Each cell is either *dead* (state 0) or *alive* (state 1). A cell's state changes with time, based on the states of its eight nearest neighbours (the *Moore neighbourhood*). Time passes in discrete intervals and the states of cells at time $t$ uniquely determine the states of cells at time $t + 1$. The initial states at time $t = 0$ are chosen by the player of the game. The initial states form a *seed pattern* that determines the course of the game, analogous to the way an organism's genome determines its phenome. The game has only one player.

The *Immigration Game* is a two-player variation of the Game of Life. It was invented by Don Woods and described in *Lifeline* by Robert Wainwright (Wainwright, 1971). The Immigration Game is similar to the Game of Life, except there are two different live states (states 1 and 2). The two live states are represented by red and blue colours. One player chooses the initial seed pattern for red and the other player chooses the pattern for blue. The two colours then compete for survival.

*Model-T* is a program that evolves populations of seed patterns for playing the Immigration Game (Turney, 2019). Natural selection in Model-T is based on fitness as determined by competition between red and blue seed patterns. The model includes asexual reproduction, sexual reproduction, and symbiosis. Model-T appears to show *open-ended evolution* (ongoing creativity) over the course of a run, although possibly the creativity may end, given a sufficiently long run.

The Game of Life belongs to a family that has 262,144 members (2 to the power of 18). The method that Woods used to transform the Game of Life into the Immigration Game extends readily to 8,192 members of the Game of Life family (2 to the power of 13). We call the original Immigration Game the *Life Immigration Game* and we call the 8,192 variations on the original game *Immigration Games*, a set that includes the original Life Immigration Game.

The 8,192 Immigration Games each have different rules. Some of these rules result in open-ended evolution over the course of a run in Model-T and other rules do not. The question that concerns us in this paper is, *what are the conditions for these rules to yield open-ended evolution in Model-T?*

The Game of Life is known to be *Turing-complete* (Rendell, 2016). That is, the rules for the Game of Life define a universal computer. In past work (Turney, 2019), it was conjectured that open-ended evolution in Model-T with the Life Immigration Game may have been due to the Turing-completeness of the Game of Life. Our core result is evidence that Turing-completeness is a *sufficient* condition for the rules of Immigration Games to yield open-ended evolution in Model-T, but not a *necessary* condition. The experiments support the hypothesis that a necessary and sufficient condition for the rules of Immigration Games to yield open-ended evolution in Model-T is that the rules allow *growth* of the seed patterns. (We give a precise, technical definition of *growth* later in the paper.)

In the following sections, we define a general family of cellular automata that includes the Game of Life (*semitotalistic rules*), and then we define the family of Immigration Games. Next, we introduce Model-T and explain how it adds evolution to Immigration Games. This is followed by a description of three different groups of Immigration Games that will serve as our experimental test cases. The core of the paper is the experiments, in which we run Model-T with a variety of Immigration Games, to investigate how different rules affect the evolution of populations of seed patterns. The paper ends with discussion of the results, future work, limitations, and the conclusion. The source code for this project is available for downloading (Turney, 2020).

## Classes of Cellular Automata

The rule for the Game of Life can be compactly represented as B3/S23, where B means *born* and S means *survives*. A cell is *born* (it switches from state 0 to state 1) when exactly three of its eight nearest neighbours are alive (they are in state 1). A cell *survives* (it stays in state 1) when it has either two or three living neighbours. Otherwise, a cell dies (it switches to state 0).

The number of living nearest neighbours for a given cell ranges from 0 to 8. The family of *semitotalistic rules* can be represented as B$x$/S$y$, where $x$ and $y$ are generated by deleting digits from the string 012345678, including deleting no digits or deleting all digits (Eppstein, 2010). This yields 2 to the power of 18 (262,144) possible semitotalistic rules. The Game of Life is the most famous member of this family.

Packard and Wolfram proposed a classification scheme that applies to semitotalistic rules for cellular automata (Packard and Wolfram, 1985). Given a random initial configuration of ones and zeros, a game eventually settles into one of four patterns: (1) the pattern of states is homogenous, (2) the pattern consists of separated groups of simple stable or periodic structures, (3) the pattern is chaotic, or (4) the pattern consists of complex localized or mobile structures.

Packard and Wolfram take the perspective of statistical physics when they consider random initial configurations, which are analogous to random distributions of molecules in a gas. Eppstein takes the perspective of an engineer, looking for ways to combine simple structures to build more complex structures with interesting behaviours (Eppstein, 2010). These kinds of complex structures would rarely occur randomly; hence the classes of Packard and Wolfram are not very useful for an engineer. Likewise, they are not useful for understanding the evolution of cellular automata by natural selection (our focus here), which can be viewed as a kind of engineering.

Eppstein proposes an alternative classification scheme (Eppstein, 2020): (1) *contraction impossible:* if a rule includes B1, any pattern expands to infinity, (2) *expansion impossible:* if a rule without B1 does not include B2 or B3, no pattern can expand, and (3) *both expansion and contraction are possible:* only when a rule does not include B1 but includes B2 or B3 can we have both expansion and contraction. For example, the Game of Life (B3/S23) falls in the third class. Eppstein expects this third class will be the most interesting class from an engineering perspective. Turing-completeness (Rendell, 2016) seems to require both expansion and contraction. We will explore Eppstein's classification later in the paper.

Thirty semitotalistic rules are known to be Turing-complete (Naszvadi, 2017). It is likely that there are many more than thirty semitotalistic Turing-complete rules, but it will probably be a long time before all 262,144 semitotalistic rules are classified as either Turing-complete or incomplete.

Inspired by Eppstein's classification scheme, we developed a simple algorithm that characterizes a rule by how often a random initial pattern *shrinks*, remains *stable*, or *grows*. The aim was not to assign each semitotalistic rule to one of these categories, but to characterize a rule by a triple of numbers, indicating how likely each category is for a given rule.

This algorithm starts with a randomly generated seed pattern, contained within a 16 × 16 box of cells. The game then runs for 200 steps. If the number of live cells in the final step is less than 80% of the number of live cells in the initial seed pattern, then the result is classified as *shrink*. If the number of live cells in the final step is more than 125% of the number of live cells in the initial seed pattern, then the result is classified as *grow* (100% / 80% = 125%). Otherwise the result is *stable*. This process is repeated 1000 times to create a triple of numbers that captures the behaviour of the rule. For example, the triple for the Game of Life is [71% *shrink*, 15% *stable*, 15% *grow*] (due to rounding, the numbers do not add to 100%). These triples can be computed rapidly (Turney, 2020).

## Immigration Games

The Life Immigration Game is almost the same as the Game of Life, except that there are two *live* states, usually represented by red and blue colours (Wainwright, 1971). The rule for updating states remains B3/S23, but there are new rules for determining colour: (1) Live cells do not change colour unless they die (*dead* is usually white). (2) When a new cell is born, it takes the colour of the majority of its neighbours.

The initial states at time $t = 0$ are chosen by the two players of the game; one player makes a red seed pattern and the other player makes a blue seed pattern. The players agree on a time limit for the game, given by a maximum value for $t$.

If states 1 and 2 were displayed with the same colour (say, black), playing the Immigration Game would appear exactly identical to playing the Game of Life. The different colours are simply a way of keeping score, to turn the Game of Life into a competitive two-player game.

The score for a colour is the number of live cells of that colour when the game ends, minus the number of live cells of that colour when the game began. If the number of live cells at the end is *less* than the number at the beginning, we assign a score of zero. The reason for subtracting the initial number of live cells from the score is to avoid giving a bias towards seed patterns with many live cells. The winner of the Life Immigration Game is the player with the highest score. (In previous work (Turney, 2019), we did not subtract the initial number of live cells from the score.)

Because the Game of Life has the rule B3/S23, there is always a clear majority, so we know what colour to assign when a cell is born. Since B3 is odd, there can be no ties. To generalize the Immigration Game to the family of semitotalistic rules, we limit the rules for birth to odd numbers. Immigration Games can be represented as B$x$/S$y$, where $x$ is generated by deleting digits from the string 1357 and $y$ is generated by deleting digits from the string 012345678. This yields 2 to the power of 13 (8,192) possible rules for Immigration Games.

We could invent a new rule to break ties for even numbers, but we prefer to keep the rules simple. To stay in the spirit of the semitotalistic rules, a new rule should preserve spatial symmetry and it should be deterministic. These requirements exclude breaking ties randomly and breaking ties with a spatially asymmetric rule. Forbidding even numbers for birth seems to be the simplest way to generalize Immigration Games.

## Model-T

Evolution by natural selection requires variation, heredity, and *differential fitness* (natural selection) (Brandon, 1996; Godfrey-Smith, 2007). In Model-T, differential fitness is calculated from one-on-one competitions in Immigration Games (Turney, 2019). An individual in the population (a seed pattern) is selected for reproduction based on how well it competes with other individuals when using the given Immigration Game rule. Model-T includes genetic mutation, asexual reproduction with genomes (seed patterns) of constant size, asexual reproduction with genomes of variable size, sexual reproduction with genetic crossover, and symbiosis by genetic fusion.

The Golly software (Trevorrow et al., 2020) is used to run Immigration Games. Variation, heredity, and differential fitness are implemented outside of Golly using algorithms written in Python (Turney, 2020). Model-T uses a steady-state model of evolution with a constant population size. In the experiments reported later in this paper, the population size is fixed at 100 individuals. New individuals are born one at a time and each new individual replaces the least fit individual in the current population. A generation is defined as the span of time over which 100 individuals are born. A run lasts for 50 generations; hence a run spans $50 \times 100 = 5,000$ births.

Model-T uses two different measures of fitness, a *relative*, *internal* measure of fitness and an *absolute*, *external* measure of fitness. Internally, Model-T calculates the fitness of an individual by having it compete with every other individual. Its internal, relative fitness is the number of games it has won divided by the total number of games it has played. The internal fitness of every individual is updated every time a new individual is born into the population.

Since every win for one individual in the population is matched by a loss for another individual, the population's average internal fitness is always 50%. By definition, it is impossible for the average internal fitness of the population as a whole to increase or decrease over time, although individuals within the population will vary in fitness over time. In general, an individual's relative fitness within a population will go down over time, as new individuals are born.

However, an external fitness measure can show that individuals, and the population as a whole, are indeed getting better at playing the Immigration Game. We measure the external fitness of an individual by having it compete with many randomly generated individuals of the same size (the same bounding box) and density (the same number of live cells). This external fitness is an absolute measure of fitness in that it is independent of the population and the generation. The external fitness measure has no impact on the evolution in Model-T. It is only used afterwards for analysis of completed runs of Model-T.

In past experiments with Model-T and the B3/S23 Life Immigration Game (Turney, 2019), when symbiosis was activated, external fitness increased steadily throughout the runs. The model demonstrated open-ended evolution for the run lengths that were used in the experiments. It is possible that evolution may be closed with very long runs, but this has not yet been observed.

Figure 1 shows a contest between two seeds, using the Life Immigration Game (B3/S23) rule. The seeds in this figure are sampled from the final generation of a run of Model-T (generation 50). Although larger seeds often win, in this case the smaller red seed pattern has won the game.

The seeds in Figure 1 are competing on a finite toroidal grid, rather than a potentially infinite planar grid. The motivation for the finite toroidal grid is to encourage competition and interaction between the two seeds by limiting the room for growth. The size of the toroid is scaled to the size of the seeds, so that larger seeds have more space to grow. The time limit for a run also increases with the size of the seeds, to allow more time for the game to unfold.

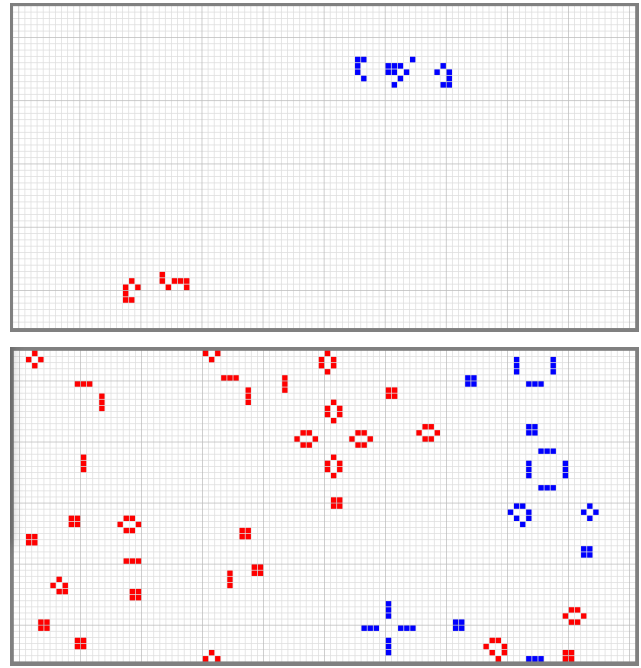

Figure 1: The first image above shows the initial state of an Immigration Game and the second image shows the final state, when the game reaches its time limit. The initial red seed is an $11 \times 5$ block of squares containing 13 live cells. The initial blue seed is a $17 \times 5$ block of squares containing 20 live cells. In the end, red won, with 121 live cells, versus 43 for blue.

Figure 1 is typical for Immigration Games that demonstrate open-ended evolution, in terms of the amount of growth from the initial seed to the final state, and in terms of the degree of interaction between the two colours. Ties are relatively rare. The outcome of a contest is often sensitive to the position and orientation of the seeds at the beginning of a game; therefore Model-T repeats games with various positions and rotations of the seeds.

## Three Groups of Immigration Rules

We previously hypothesized that the apparent open-endedness of the Life Immigration Game (B3/S23) with Model-T may have been due to the fact that Life is Turing-complete (Turney, 2019). To test this hypothesis, we created three groups of Immigration Games. Each group was sampled from the 8,192 possible Immigration Games using different criteria, with the goal of creating three diverse samples.

Group 1 is a subset of the thirty semitotalistic rules that are known to be Turing-complete (Naszvadi, 2017). Thirteen of the thirty rules belong to the 8,192 possible rules for Immigration

Games; that is, thirteen have only odd numbers for the *born* parts of their rules. Group 1 consists of these thirteen rules, including the Life Immigration Game rule, B3/S23.

Group 2 is a random sample (using the Python random number generator) from the set of 8,192 possible rules for Immigration Games. The size of the random sample was limited to thirteen rules in order to match the size of Group 1. Any semitotalistic rule with odd numbers for *born* was possible for Group 2.

Group 3 was based on the intersection of Eppstein's three criteria with the rules for Immigration Games (Eppstein, 2020). Rules were selected from the set of Immigration Games rules subject to the requirement that the *born* part of the rule could not contain B1 but must contain B3. This follows from Eppstein's third condition (both expansion and contraction are possible only when a rule does not include B1 but includes B2 or B3) combined with the requirement that Immigration Games can only have odd *born* numbers (which excludes B2). Such rules can be represented as B3$x$/S$y$, where $x$ is generated by deleting digits from the string 57 and $y$ is generated by deleting digits from the string 012345678. This yields 2 to the power of 11 (2,048) possible rules. The random sample (using the Python random number generator) was limited to thirteen rules, in order to match the sizes of Groups 1 and 2.

## Evolution of the Three Groups

Table 1 summarizes the results of the experiments, running Model-T with the three groups of Immigration Games. The first column identifies the three different groups of Immigration Games. The second column specifies the rule for each Immigration Game. Note that the first rule in the table is the rule for the Game of Life.

The third column in Table 1 classifies the run as *open*, *closed*, or *none*. This classification was made by graphing the external fitness of the population over the course of a run. When the external fitness increased at a steady pace throughout a run, the run was labeled as being an instance of open-ended evolution (*open*). When the external fitness increased at the start of a run but eventually stopped improving, the run was labeled as being an instance of *closed* evolution. When the external fitness was random (50%) throughout the whole course of a run, the run was labeled as *none* (no sign of any evolution in external fitness).

The fourth column in Table 1 gives the external fitness in the final generation (generation 50). The external fitness is measured using a sample of the top half of the population (the top 50 individuals out of a population of 100 individuals). The population is ranked by internal fitness in order to determine the top individuals.

Columns five, six, and seven in Table 1 provide the [*shrink*, *stable*, *grow*] triples for each rule. The interesting result here is that a rule is an instance of open-ended evolution (Evolution = Open in Table 1) if and only if its growth is greater than zero (% Grow > 0 in Table 1). This means that we can predict whether a rule will be open-ended without running Model-T and observing the external fitness. Instead, we can simply calculate the rule's [*shrink*, *stable*, *grow*] triple. Running Model-T for 50 generations with a population of 100 individuals can take about two weeks, whereas calculating a rule's [*shrink*, *stable*, *grow*] triple takes a couple of minutes.

Since we have defined internal fitness based on growth (see the section above, titled *Immigration Games*), it is not surprising that open-ended evolution in Model-T requires growth, but it is surprising (to us) that growth is *all* that is required. We expected that there would be an additional requirement for a certain degree of complexity in the rules, such as the complexity indicated by Turing-completeness.

We do not have a proof that nonzero growth predicts open-ended evolution; we only have the evidence of the experiments. It is possible that the apparently open-ended evolution will end, given a sufficiently long run of Model-T. It is also possible that a rule might be capable of growth, but only with certain rare initial states, which might never occur in the random configurations used to calculate a rule's [*shrink*, *stable*, *grow*] triple. However, nonzero growth predicts the presence of open-ended evolution for the 39 rules in Table 1 with no errors, which suggests that it is a reliable indicator, even if it is theoretically possible that it does not always make the correct prediction.

In the third column of Table 1, both *closed* and *none* correspond to a zero value in column seven (*grow*). These two different kinds of failure to evolve cannot be distinguished using [*shrink*, *stable*, *grow*] triples. Our main interest here is in open-ended evolution, so the distinction between *open* versus *not open* is more important to us than the distinction between *closed* and *none*. Although we have no specific need to distinguish *closed* and *none*, it might be interesting for future work to see whether they can be distinguished without the expense of running Model-T.

There are some rules with an external fitness near 50.0%, yet we have classified them as *open*, not *none*. Three rules in Group 2 deserve some discussion: B1357/S178 (classified as *open* with an external fitness of 50.2%), B1357/S468 (*open*, fitness 51.0%), and B157/S048 (*open*, fitness 52.0%). All three have an external fitness near 50.0%, and so it seems perhaps they should be classified as not showing evolution (*none* in the third column), since their external fitness is nearly random (nearly 50.0%). However, visual inspection shows that the individuals in the populations are interacting and competing vigorously. Their behaviour is quite different from the behaviour when rules have an external fitness of exactly 50.0%. We believe that the external fitness of these three rules will eventually rise higher, but it will require longer runs (more generations) and larger populations, in order to give evolution more time to increase the fitness.

There are also some rules with an external fitness near 100.0%, yet we have classified them as *open*, not *closed*. For example, B3/S2345678 in Group 3 has an external fitness of 97.5%, B37/S34567 in Group 3 has an external fitness of 96.4%, and B37/S237 in Group 1 has an external fitness of 96.0%. However, when we look at the graphs of the external fitness of the population over the course of a run, these three rules show a slow, steady growth in fitness, whereas the *closed* rules reach an external fitness near 100.0% much faster, around generation 25.

It is interesting to see that the [*shrink*, *stable*, *grow*] triples for Group 1 (the Turing-complete rules) are not like the triples for Groups 2 and 3. The triples for Groups 2 and 3 are polarized, with extreme values for *shrink* or *grow* and values near zero for *stable*. Group 1 has less polarization and *stable* ranges from 0% to 23%. As Eppstein suggests, these more balanced triples may be an indicator of Turing-completeness (Eppstein, 2010).

| Group | Rule | Evolution | % Fitness | % Shrink | % Stable | % Grow |
|---|---|---|---|---|---|---|
| 1 | B3/S23 | Open | 90.0 | 71 | 15 | 15 |
| 1 | B3/S236 | Open | 84.5 | 5 | 0 | 95 |
| 1 | B3/S2367 | Open | 83.4 | 4 | 0 | 96 |
| 1 | B3/S23678 | Open | 82.8 | 3 | 0 | 97 |
| 1 | B3/S2368 | Open | 89.0 | 5 | 0 | 95 |
| 1 | B3/S237 | Open | 90.5 | 44 | 23 | 33 |
| 1 | B3/S2378 | Open | 84.9 | 44 | 18 | 38 |
| 1 | B3/S238 | Open | 89.8 | 66 | 15 | 20 |
| 1 | B35/S236 | Open | 79.2 | 3 | 0 | 97 |
| 1 | B37/S23 | Open | 89.7 | 59 | 18 | 23 |
| 1 | B37/S236 | Open | 82.2 | 4 | 0 | 95 |
| 1 | B37/S237 | Open | 96.0 | 35 | 18 | 48 |
| 1 | B37/S238 | Open | 87.5 | 54 | 20 | 27 |
| 2 | B13/S234567 | Open | 76.0 | 0 | 0 | 100 |
| 2 | B13/S3456 | Open | 70.8 | 0 | 0 | 100 |
| 2 | B1357/S178 | Open | 50.2 | 0 | 0 | 100 |
| 2 | B1357/S468 | Open | 51.0 | 0 | 0 | 100 |
| 2 | B157/S048 | Open | 52.0 | 0 | 0 | 100 |
| 2 | B3/S2348 | Open | 93.2 | 2 | 0 | 98 |
| **2** | **B35/S0257** | **Closed** | **99.6** | **100** | **0** | **0** |
| **2** | **B35/S247** | **None** | **50.0** | **100** | **0** | **0** |
| **2** | **B357/S078** | **None** | **50.0** | **100** | **0** | **0** |
| 2 | B357/S12357 | Open | 66.0 | 0 | 0 | 100 |
| **2** | **B37/S03578** | **None** | **50.0** | **100** | **0** | **0** |
| **2** | **B57/S13478** | **None** | **50.0** | **100** | **0** | **0** |
| **2** | **B7/S02348** | **None** | **50.0** | **100** | **0** | **0** |
| 3 | B3/S012678 | Open | 71.6 | 0 | 0 | 100 |
| 3 | B3/S013468 | Open | 82.2 | 1 | 1 | 99 |
| **3** | **B3/S025** | **Closed** | **99.7** | **100** | **0** | **0** |
| 3 | B3/S2345678 | Open | 97.5 | 1 | 0 | 99 |
| **3** | **B3/S457** | **None** | **50.0** | **100** | **0** | **0** |
| **3** | **B35/S0256** | **Closed** | **100.0** | **100** | **0** | **0** |
| 3 | B35/S13456 | Open | 92.0 | 1 | 0 | 99 |
| 3 | B35/S134567 | Open | 85.4 | 1 | 0 | 99 |
| 3 | B357/S023567 | Open | 81.5 | 1 | 0 | 99 |
| 3 | B357/S1247 | Open | 83.7 | 1 | 0 | 99 |
| **3** | **B357/S8** | **None** | **50.0** | **100** | **0** | **0** |
| 3 | B37/S023456 | Open | 88.8 | 1 | 0 | 99 |
| 3 | B37/S34567 | Open | 96.4 | 6 | 1 | 93 |

Table 1: The three groups of rules and their characteristics. All of the rules in Group 1 (the Turing-complete rules) show open-ended evolution. Groups 2 and 3 are mostly open but sometimes show closed evolution or no evolution (*none*). Evolution is open-ended if and only if the percentage of growth is greater than zero. Closed evolution is associated with fitness near 100% in the final generation. When there is no evolution, fitness is near 50% throughout the generations.

Figure 2 is a graph of the average external fitness for the three groups over the course of a run. The average external fitness of Group 1 (the Turing-complete rules) rises steadily throughout a run of Model-T. Group 3 (Eppstein's rules) performs almost as well as Group 1, but the fitness curve flattens out around generation 25. The two *closed* rules in Group 3 have reached their maximum values by generation 25. The two rules with no evolution (*none*) also bring down the average fitness of Group 3 throughout the entire run. The external fitness of Group 2 (randomly selected Immigration Games) is quite low, compared to Groups 1 and 3. The fitness hits a peak around generation 10 and then flattens out. The five rules with no evolution (*none*) and the one *closed* rule drag down the average fitness of Group 2.

Figure 3 shows the average external fitness when the 39 rules are grouped according to their evolutionary behaviours. The *closed* group consists of the three rules for which the external fitness increased at the start of a run but eventually stopped improving. This is clear in Figure 3, where the graph for *closed* rises quickly and then flattens out around generation 25. The seven rules with no evolution (*none*) have an external fitness of 50.0% (random) through all of the generations. The remaining 29 *open* rules show slow and steady progress.

In Figure 3, it might seem that the three *closed* rules are performing well, since they reach an external fitness near 100.0% (99.6% for B35/S0257 in Group 2, 99.7% for B3/S025 in Group 3, and 100.0% for B35/S0256 in Group 3). However, we are interested in open-ended evolution, and thus a fitness of 100.0% is not good, because it means the evolution has reached a maximum; there is no room for further improvement.

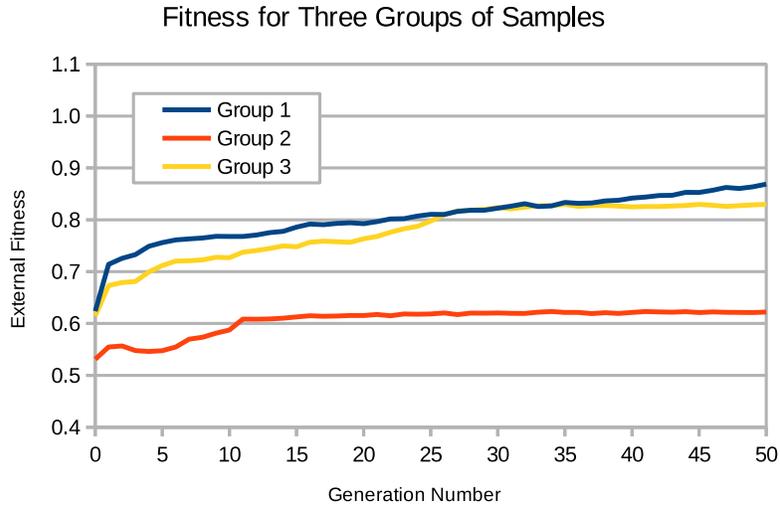

Figure 2: The average external fitness for each of the three groups of rules. The rules in Group 1 (the Turing-complete rules) show a steady increase in external fitness over the generations. The rules in Group 3 (Eppstein's rules) hit a plateau in the later generations. The rules in Group 2 (randomly selected rules) do not fare as well as rules in the other two groups.

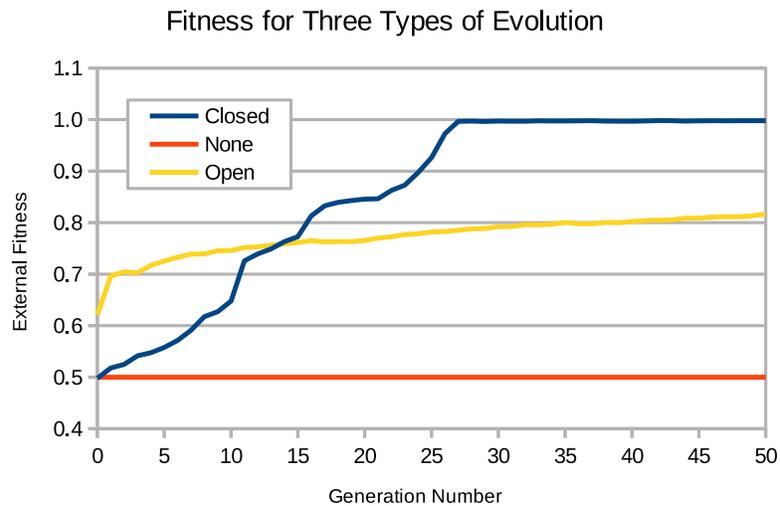

Figure 3: The average external fitness for rules grouped by their behaviours: *closed* (evolution proceeds for several generations and then stops), *none* (fitness is entirely random for all generations), or *open* (fitness indicates open-ended evolution).

Recall how external fitness is defined (in the section titled *Model-T*): We measure the external fitness of an individual by having it compete with many randomly generated individuals of the same size (the same bounding box) and density (the same number of live cells). If an individual in the population wins all of the games against these randomly generated competitors, it suggests that there is no way to improve upon that individual. Evolution is stuck at a maximum in the fitness landscape. The flat blue line in Figure 3 means that evolution has stalled. Only the yellow line shows ongoing improvement.

It might seem that the yellow line (*open*) in Figure 3 must eventually reach 100.0% if we run Model-T for a sufficient number of generations, but this is not necessarily true. When the population's average external fitness gets very close to

100.0%, we can reduce the level of randomness in the randomly generated competitors. This will make the competition more difficult and it will shift all of the fitness curves downwards, making more room for improvement. It seems possible for external fitness to improve without bound, if we periodically make the fitness test more difficult, as the population evolves increasingly better strategies for playing Immigration Games.

Note that the internal, relative fitness measure needs no such adjustment as the population evolves, since the fitness is always relative to the current population. This is the reason for having both an internal, relative fitness measure (for evolutionary selection of individuals) and an external, absolute fitness measure (for making comparisons among individuals across different populations).

## Discussion of Results

In the context of evolution by natural selection in Immigration Games with Model-T, focusing on the conditions that the rules must satisfy, the evidence in Table 1 indicates that Turing-completeness is a sufficient condition for open-ended evolution to occur, but it is not a necessary condition. On the other hand, growth, as measured by [*shrink*, *stable*, *grow*] triples, is both necessary and sufficient for open-ended evolution.

The experiments support these claims, but we do not have a mathematical proof of the claims. For the 39 rules that we have examined, open-ended evolution is perfectly predicted by nonzero growth, but it is possible that there are exceptions somewhere in the family of 8,192 rules for Immigration Games.

Group 2 is a random sample from the set of 8,192 possible rules for Immigration Games. We can see from Table 1 that seven of the thirteen rules in Group 2 demonstrate open-ended evolution. Extrapolating from this sample, it seems there should be at least 4,000 rules for Immigration Games that are open-ended (8,192 $\times$ 7/13 = 4,411). This is a large space to explore for those who are interested in open-ended evolution.

It was our hope that the family of Immigration Games would be a rich space for exploring open-ended evolution, but the hypothesis that open-ended evolution requires Turing-completeness was somewhat discouraging for this hope, since only thirteen of the 8,192 rules for Immigration Games are known to be Turing-complete. Hence it is encouraging to discover that many of the 8,192 rules for Immigration Games are likely to support open-ended evolution.

## Future Work and Limitations

The support for the claims made here is based on 39 cases. We would like more cases, but one run of Model-T can take about two weeks, depending on the computer hardware and the Model-T parameter settings. Of course, with many processors, many runs can be executed in parallel, so this is an issue that could be addressed with more resources.

It might be possible to find a theorem that covers all 8,192 rules, proving that there is open-ended evolution in Immigration Games if and only if there is growth, but we do not yet know how to approach this proof. There might be some connections to other problems in computer science. For example, *general recursive functions* are Turing-complete and *primitive recursive functions* are a strict subset of general recursive functions (Soare, 1996). The Game of Life is known to be Turing-complete (Rendell, 2016) but the rules in Group 2 that contain B1 might not be Turing-complete (Eppstein, 2020). This suggests that some open-ended Immigration Games may correspond to general recursive functions and others may correspond to primitive recursive functions. Distinguishing different computational classes of open-ended Immigration Games could make the theorem easier to prove (the divide-and-conquer approach to theorems).

In future work, we plan to modify Model-T to explore the coevolution of genes and genotype-phenotype maps. The initial seed pattern in an Immigration Game is analogous to a gene and the unfolding of the initial pattern as the game runs is analogous to the development of the phenotype from the genotype (McCaskill and Packard, 2019). The map from genotype to phenotype is determined by the chosen Immigration Game rule. Model-T currently uses natural selection to evolve new genes and introduce them into the population, but Model-T does not currently modify the Immigration Game rule during a run; the rule is determined at the start of a run by the user. Our plan is to modify Model-T so that the Immigration Game rule (the genotype-phenotype map) is also subject to mutation and selection, just as the seed pattern (the gene) is subject to mutation and selection.

## Conclusion

In this paper, we raised the question, *what are the conditions on rules for Immigration Games to exhibit open-ended evolution?* An earlier hypothesis that Turing-completeness was necessary (Turney, 2019) now seems to be incorrect, although Turing-completeness appears to be a sufficient condition. The experiments presented here support the claim that *growth* (as defined in a technical sense in this paper) is a necessary and sufficient condition for open-ended evolution in Immigration Games. This is a welcome result, because it suggests that there are thousands of Immigration Games that can be used to study open-ended evolution.

## Acknowledgements

Thanks to Martin Brooks and Daniel Lemire for helpful discussions about this work.